\documentclass[10pt,twocolumn,letterpaper]{article}

\usepackage{iccv}

\usepackage{times}
\usepackage{epsfig}
\usepackage{graphicx}
\usepackage{amsmath}
\usepackage{amssymb}
\usepackage{mcode}
\usepackage{caption}

\usepackage{subfigure}

\usepackage{makecell}
\usepackage{booktabs}
\usepackage{multirow}
\usepackage{siunitx}

\usepackage[T1]{fontenc}
\usepackage{fix-cm}

\newcolumntype{?}{!{\vrule width 1pt}}
\newcolumntype{C}[1]{>{\centering}m{#1}}

\newcolumntype{X}{@{\hskip\tabcolsep\vrule width 1.5pt\hskip\tabcolsep}}

\newcommand{\myfigurethreecol}[1]{
\begin{minipage}[b]{.14\textwidth}
\includegraphics[width=1.1\linewidth]{#1}
\end{minipage}
}

\newcommand{\myfigurethreecolcaption}[2]{
\begin{minipage}[b]{.14\textwidth}
\includegraphics[width=1.1\linewidth]{#1}
\caption{{\small {#2}}}
\end{minipage}
}

\usepackage[pagebackref=true,breaklinks=true,letterpaper=true,colorlinks,bookmarks=false]{hyperref}

\iccvfinalcopy 

\def\iccvPaperID{882} 

\ificcvfinal\fi
\begin{document}

\title{Unsupervised Learning of Important Objects from First-Person Videos}

\author{Gedas Bertasius$^{1}$, Hyun Soo Park$^2$, Stella X. Yu$^3$, Jianbo Shi$^1$\\
$^1$University of Pennsylvania, $^2$University of Minnesota, $^3$UC Berkeley ICSI\\
{\tt\small \{gberta,jshi\}@seas.upenn.edu} \ \ \ {\tt\small hspark@umn.edu}  \ \ \ {\tt\small stella.yu@berkeley.edu}
}


\maketitle

\begin{abstract}

A first-person camera, placed at a person's head, captures, which objects are important to the camera wearer. Most prior methods for this task learn to detect such important objects from the manually labeled first-person data in a supervised fashion. However, important objects are strongly related to the camera wearer's internal state such as his intentions and attention, and thus, only the person wearing the camera can provide the importance labels. Such a constraint makes the annotation process costly and limited in scalability.

In this work, we show that we can detect important objects in first-person images without the supervision by the camera wearer or even third-person labelers. We formulate an important detection problem as an interplay between the 1) segmentation and 2) recognition agents. The segmentation agent first proposes a possible important object segmentation mask for each image, and then feeds it to the recognition agent, which learns to predict an important object mask using visual semantics and spatial features.  

We implement such an interplay between both agents via an alternating cross-pathway supervision scheme inside our proposed Visual-Spatial Network (VSN). Our VSN consists of spatial (``where'') and visual (``what'') pathways, one of which learns common visual semantics while the other focuses on the spatial location cues. Our unsupervised learning is accomplished via a cross-pathway supervision, where one pathway feeds its predictions to a segmentation agent, which proposes a candidate important object segmentation mask that is then used by the other pathway as a supervisory signal. We show our method's success on two different important object datasets, where our method achieves similar or better results as the supervised methods.

\end{abstract}

\vspace{-0.45cm}

\section{Introduction}

A question ``what is where?'' attempts to delineate a picture as a spatial arrangement of objects rather than a collection of unordered visual words, which inspires core computer vision tasks such as recognition, segmentation, and 3D reconstruction. This spatial arrangement encodes not only the physical relationship between objects in front of the camera but also the interactions with {\em the photographer standing behind the camera}\footnote{Figure-ground segmentation, and saliency detection are a line of work that addresses the relationship with the photographer.}. A picture is always taken by a photographer reflecting what is important to her/him, which provides a strong cue to infer the internal states such as his/her intent, attention, and emotion. In particular, first-person videos capture unscripted interactions with scenes suggesting that the spatial layout is arranged such that the objects can afford the associated actions, e.g., a cup appears to be held by right hand from the holder's point of view.

\captionsetup{labelformat=empty}
\captionsetup[figure]{skip=5pt}

\begin{figure}
\centering

%
%

\includegraphics[width=1\linewidth]{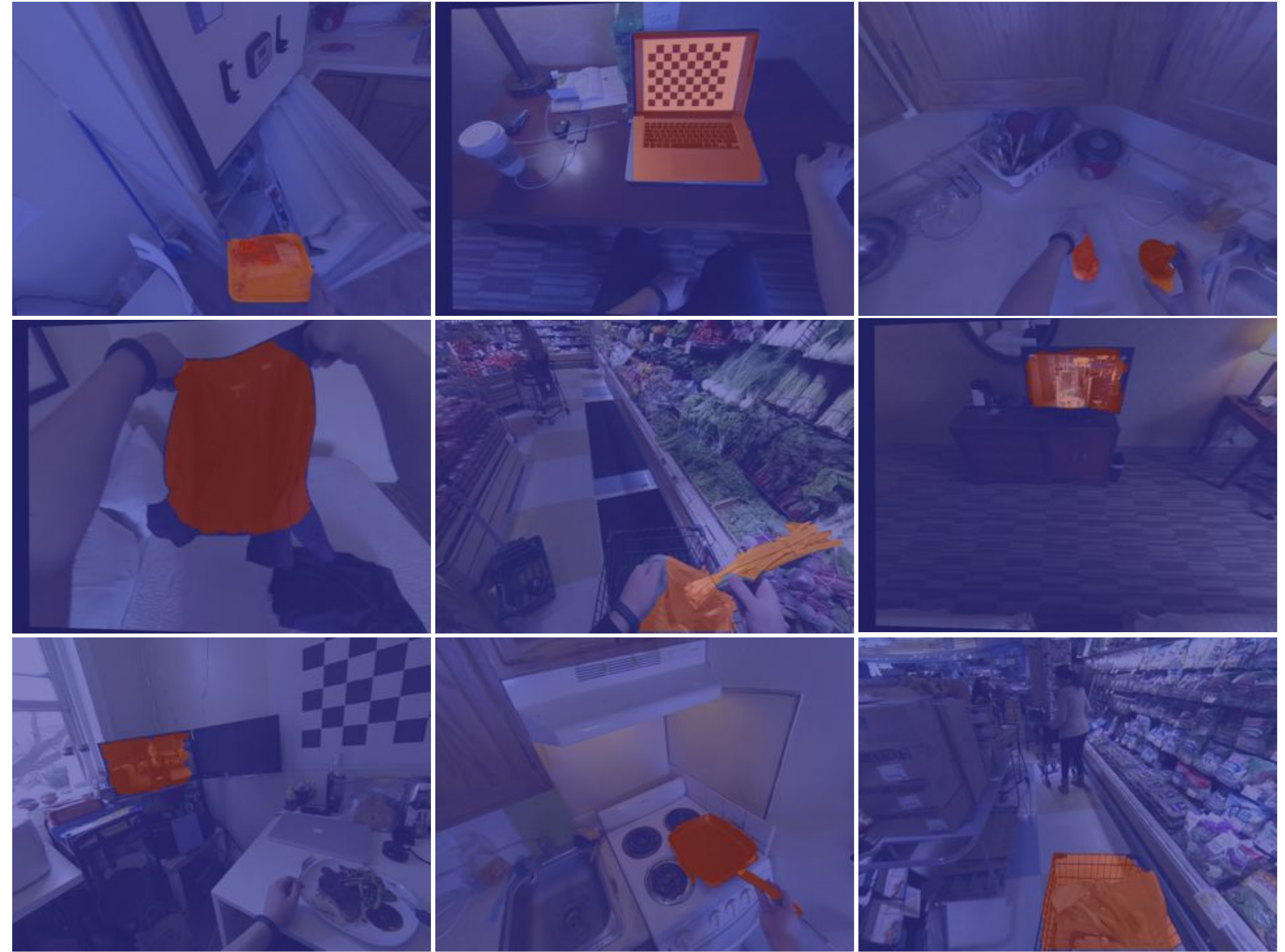}

\captionsetup{labelformat=default}
   \setcounter{figure}{0}
    \caption{Given an \textbf{unlabeled} set of first-person images our goal is to find all objects that are important to the camera wearer. Unlike most prior methods, we do so without using ground truth importance labels.\vspace{-0.5cm}}
    \label{task_fig}
\end{figure}

\captionsetup{labelformat=default}




In this paper, we aim to detect objects that are important to the photographer from a first-person video. Since importance is a subjective matter, the photographer is the only one who can identify an important object. However, we conjecture that it is possible to detect important objects without the supervision by the photographer or even third-person labelers because an important object exhibits common visual semantics (what it looks like) and a spatial layout (where it is in the first-person image).



To achieve this goal, we formulate an important object detection task as an interaction between the 1) segmentation and 2) recognition agents. Initially, the segmentation agent generates a candidate important object mask for each image, and relays this mask to the recognition agent, which then tries to learn a classifier to predict  such an important object mask using visual semantics and spatial cues.


Our segmentation agent is implemented using an MCG projection scheme, which employs the samples generated from an unsupervised segmentation method~\cite{APBMM2014} to propose important object segmentation masks to the recognition agent. Our recognition agent is implemented using the visual (``what'') and spatial (``where'') pathways of our proposed Visual-Spatial Network (VSN),  each of which learns to predict important object masks by asking questions  ``what an important object looks like?'' and ``where an important object is in the first-person image?''. We design these pathways using a fully convolutional network (FCN) while also embedding a location dependent layer in the spatial pathway to learn the first-person spatial location prior. 


Our VSN then learns to detect important objects without using manually annotated importance labels. We do so via an alternating cross-pathway supervision, in a synergistic interplay between visual (``what'') and spatial (``where'') pathways, and a segmentation agent. Each pathway's output is provided to a segmentation agent, which first generates a possible important object segmentation mask and then relays it to the other pathway to be used as a supervisory signal. The supervision proceeds in such an alternating fashion as each pathway improves each other, and as the segmentation agent becomes better as well.



\textbf{Why Unsupervised Learning?}  Building a framework that can learn without manually collected labels is particularly essential for first-person important object detection because the annotation task is not scalable at all unlike object detection/segmentation~\cite{imagenet_cvpr09,502} where a consensus of third parties from crowdsourcing mechanism can be used. In the important object detection task, only the camera wearer can perform the annotation task by looking back on his/her past experiences. Prior methods~\cite{Li_2015_CVPR} have used a wearable gaze tracker to label the camera wearer's visual attention. However, gaze tracker is invasive and the data that it captures has no notion of objects. Instead, our paper addresses these issues via an unsupervised alternating cross-pathway learning scheme, which allows our method to achieve similar or even better results as the supervised methods do.

\section{Related Work}


\textbf{Important Object Detection in First-Person.} There have been a number of first-person methods that explored important object detection task either as a main task~\cite{DBLP:journals/corr/BertasiusPYS16,conf/cvpr/RenG10,conf/cvpr/FathiRR11}, or as an auxiliary task for an activity recognition~\cite{PirsiavashR_CVPR_2012_1,Li_2015_CVPR,ma2016going,Fathi:2011:UEA:2355573.2356302} or video summarization~\cite{DBLP:journals/ijcv/LeeG15,Lu:2013:SSE:2514950.2516026}. The work in~\cite{DBLP:journals/ijcv/LeeG15,conf/cvpr/FathiRR11,Li_2015_CVPR,PirsiavashR_CVPR_2012_1} employ hand-crafted appearance features, egocentric and optical flow features to describe a first-person image, and then train a discriminative classifier to detect the regions that correspond to the important objects. The more recent work~\cite{ma2016going,DBLP:journals/corr/BertasiusPYS16} use FCNs~\cite{DBLP:journals/corr/LongSD14} to predict important objects end-to-end. Whereas the method in~\cite{DBLP:journals/corr/BertasiusPYS16} employs a two stream visual appearance and 3D network, the work in~\cite{ma2016going} exploits the connection between the activities and objects and proposes a two stream appearance and optical flow network with a multi-loss objective function. 

All of these methods use manually annotated important object labels, which may be costly and difficult to obtain. Our approach, on the other hand, introduces a new unsupervised learning scheme that allows us to learn important objects without manually labeled importance annotations.



\textbf{Training FCNs with Weakly-Labeled Data.}  Recently, there have been several deep learning approaches that proposed learning with weakly labeled or unlabeled datasets~\cite{Li_2016_CVPR,DBLP:journals/corr/DaiH015,Bearman16,papandreou15weak,xu_cvpr2015,pinheiro:2015a,Lin_2016_CVPR,pathakICLR15,pathakICCV15ccnn} . Due to the high cost of obtaining per-pixel labels, this has been a particularly relevant problem for semantic segmentation. 

The weakest form of supervision for semantic segmentation includes image-level labels, which were used to train FCNs in several prior approaches~\cite{pinheiro:2015a,pathakICLR15,pathakICCV15ccnn,papandreou15weak}. Some recent work~\cite{Bearman16} used point supervision, which requires almost as much effort as the image-level labels but also provides some spatial information. Several approaches employed free form squiggles as a supervisory signal~\cite{xu_cvpr2015,Lin_2016_CVPR} which provides even more information, and are still easy enough to annotate. Furthermore, several approaches utilized bounding box level annotations for FCN training~\cite{papandreou15weak,DBLP:journals/corr/DaiH015}. Finally, recent work achieved excellent edge detection results without using any annotations at all~\cite{Li_2016_CVPR}.

In comparison to prior work, which focuses on the third-person data, our method focuses on the first-person data. Unlike third-person object detection/segmentation tasks where annotations can be obtained via a crowdsourcing mechanism, important object detection task requires the camera wearer to provide the labels, which severely limits its scalability. Due to such a constraint, an unsupervised learning framework is particularly important for the important object detection task in the first-person setting.

\begin{figure*}
\begin{center}
   \includegraphics[width=0.95\linewidth]{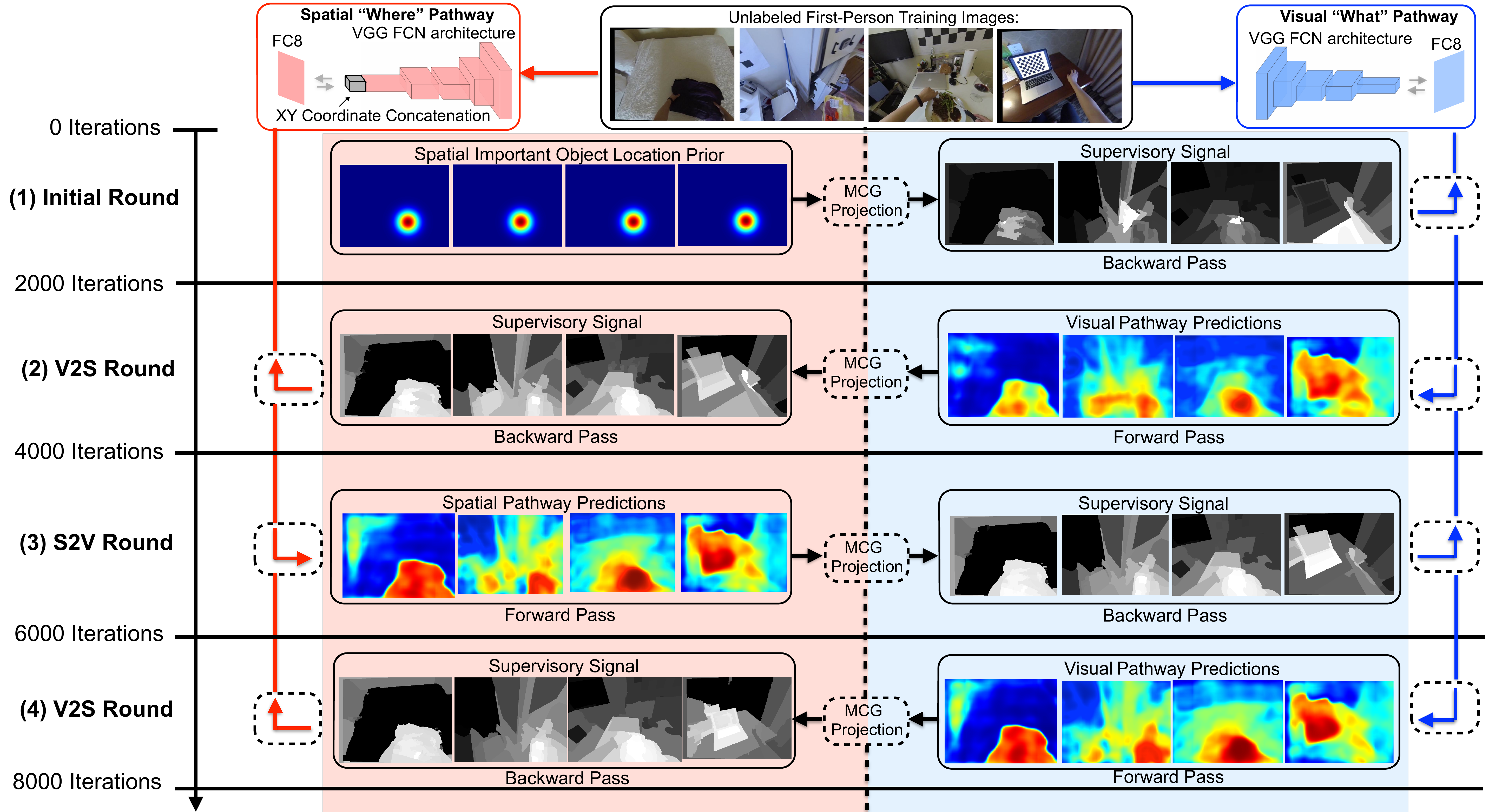}
\end{center}
\caption{ We implement an interplay between the segmentation and recognition agents via an alternating cross-pathway supervision scheme inside our proposed Visual-Spatial Network (VSN).  Our VSN consists of the 1) visual (``what'') and 2) spatial (``where'') pathways, which both act as recognition agents. In between these two pathways, the VSN uses an MCG projection scheme, which acts as a segmentation agent. Then, given a set of \textbf{unlabeled} first-person training images, we first guess ``where'' an important object is in the first-person image and use an MCG projection scheme to propose important object segmentation masks. These masks are then used a supervisory signal to train a visual pathway such that it would learn ``what'' an important object looks like. Then, in the V2S round, the predictions from the visual pathway are passed through the MCG projection, and transfered to the spatial pathway. The spatial pathway then learns ``where'' an important object is in the first-person image. Such an alternating cross-pathway supervision scheme is repeated for several rounds.\vspace{-0.5cm}}
\label{fig:arch}
\end{figure*}

\section{Approach Motivation}
\label{learning_sec}

Our goal is to 1) recognize and 2) segment important objects from a first-person image in an unsupervised setting. Thus, we want our method to have two key properties: 1) it needs to segment the important objects from the background based on the low-level grouping cues and 2) it needs to be discriminative, i.e, recognize objects that are important and ignore all the irrelevant objects. 

To achieve these goals, we frame an important object prediction task as an interplay between the 1) recognition and 2) segmentation agents, where a segmentation agent first proposes a possible important object mask, which a recognition agent then uses as a supervisory signal to learn an important object classifier based on visual (``what'') and spatial (``where'') cues.

The main challenge of our unsupervised learning framework is to prevent overfitting of either a segmentation or a recognition agent.   If the segmentation agent proposes too many different segments, the recognition agent will not learn a concept of important objects (particularly if these segments are not recurring). On the other hand, if the recognition agent narrowly focuses on predicting one type of object, or an object that appears at a particular location, it will not generalize across all images. We address the first issue by feeding the predictions from the recognition agent to the segmentation agent, so that the target segmentations would consistently improve as the recognition agent gets better. To tackle the second issue,  we force the recognition agent to learn a diverse model by making it focus on visual (``what'') and spatial (``where'') cues in an alternating fashion.


We now provide more details related to the 1) segmentation and 2) recognition agents that we want to use for our unsupervised learning task.

\subsection{Segmentation Agent} 




The goal of a segmentation agent is to propose segmentation masks of the important objects, which could then be used by a recognition agent as a supervisory signal. We implement such a segmentation agent via our introduced MCG projection scheme. We define MCG projection as a function $h(A,R)$ that takes two inputs: 1) a coarse per-pixel important object mask prediction $A$, and 2) a set of regions $R$ obtained from a segmentation method MCG~\cite{APBMM2014}. The output $h(A,R)$ then captures an important object segmentation mask proposed by a segmentation agent.

We first run an MCG~\cite{APBMM2014} segmentation algorithm, which segments a given image into regions $R$. Then, for every MCG region $R$, we compute the mean value of all values in $A$ that fall in the region $R$, and assign that value to the entire \textbf{region} $R$. Since MCG regions overlap with each other, the pixels belonging to multiple overlapping regions, get assigned multiple values (from each region they belong to). To assign a single value to a given \textbf{pixel}, we perform max-pooling, over the values of that pixel in each of the regions that contains that \textbf{pixel}. This then produces a candidate important object segmentation mask.

 
 


\subsection{Recognition Agent: Motivation}

To build a recognition agent that is discriminative, and yet generalizable, we focus on two distinct aspects of an important object prediction task: the ``what'' (what does an important object look like?) and the ``where'' (where does an important object appear in the first-person image?).

\textbf{The Visual Cues (What it looks like?)} A natural way to predict important objects is by learning ``what''  they look like. Such learned visual appearance cues can then be used to predict important objects in an image. This is exactly what is done by the supervised methods, which use the ground-truth data to learn the visual characteristics of ``what'' a prototypical important object looks like in a first-person image. However, in the context of our problem, we do not have access to such ground-truth data. Thus, the key question becomes whether we can learn to detect important objects despite not knowing ``what'' they look like beforehand?


\textbf{The Spatial Cues (Where it is?)} We conjecture that important objects are spatially arranged in the first-person image to afford the camera wearer's interactions with those objects. In other words,  by performing activities, and looking at things, the camera wearer is implicitly labeling what is important to him, which is also captured in a first-person image. For instance, a cup often appears at the \textit{bottom right} of a first-person image, because most people look down at it and also hold it with their right hand. 

Thus, since 1) people typically look down at an object, with which they interact, and 2) since most people are right-handed, we conjecture that many important objects appear at the \textit{bottom-right} of a first-person image, which we guess to be at $(x,y)$ location $(0.6W,0.75H)$, where $W$ and $H$ denote the width and height of the first-person image. We refer to this location as a spatial important object location prior.

Since we do not have ground truth labels, we cannot directly supervise our network by telling it ``what'' an important object looks like. However, we can tell the network ``where'' we think an important object is such that the network would learn the visual appearance cues necessary to recognize ``what'' appears at that location. In the best case, there will be a true important object at our specified location, and the network will then learn ``what'' that important object looks like. Otherwise, if our guess is incorrect, the network will try to learn a meaningless pattern of ``what'' something that is \textbf{not} an important object looks like. If we make enough correct guesses of ``where'' the true important objects are, our network will learn ``what'' important objects look like without ever using ground truth importance labels.

\section{Visual-Spatial Network}


To holistically integrate both segmentation and recognition agents, we introduce a Visual-Spatial Network (VSN) that learns to detect important objects from unlabeled first-person data. Our network consists of the 1) visual (``what'') and 2) spatial (``where'') pathways, which act as recognition agents. In between these two pathways, the VSN employs an MCG projection scheme, which acts as a segmentation agent.

During training, we first use an MCG projection to propose a candidate important object segmentation mask, which is then used by the visual ``what'' pathway as a supervisory signal.  Then, the predictions from the visual pathway are used by the segmentation agent to generate an improved important object segmentation mask, which is used as a supervisory signal by the spatial ``where'' pathway. Such a supervision scheme between the two pathways proceeds in an alternating fashion, allowing each pathway to improve each other, while the segmentation agent also improves. We refer to such a learning scheme as a cross-pathway supervision, which we illustrate in Fig.~\ref{fig:arch}. 

\subsection{Visual ``What'' Pathway}
\label{app}



The visual pathway of our VSN is based on a fully convolutional VGG architecture~\cite{Simonyan14c}, which is pretrained for the segmentation task on Pascal VOC dataset with $20$ distinct classes such as airplane, bus, cow, etc. We note that the classes in Pascal VOC dataset are quite different compared to the important object classes in the datasets that we use for our experiments. For instance, Pascal VOC segmentation dataset does not include annotations for classes such as food package, knife, suitcase, sweater, pizza and many more object classes. In the experimental section, we also verify this claim by showing that the VGG FCN~\cite{Simonyan14c} that was pretrained for the Pascal VOC semantic segmentation task alone produces poor important object detection results. 

We want to make it clear that we do not claim that our method does not use any annotations at all. Our main claim is that we can learn to detect important objects in first-person images without manually annotated first-person importance labels. Our network still needs a general visual recognition capability to differentiate between various visual appearance cues. Otherwise, due to a noisy supervisory signals that we use to train each pathway, our network would struggle to learn the visual cues that are indicative of true important objects. 


\subsection{Spatial ``Where'' Pathway}
\label{spatial}


The spatial pathway is also based on the pretrained VGG FCN~\cite{Simonyan14c}. However, unlike the visual pathway, the spatial pathway incorporates a two-channel grid of normalized X and Y coordinates that correspond to every pixel in the first-person image. These $X,Y$ coordinate meshgrids could be obtained by calling a matlab command \mcode{[X,Y]=meshgrid(1:W,1:H)}, where $W,H$ are the width and height of an image respectively. We then use a bilinear interpolation to downsample these grids 8 times and concatenate them to the visual \textit{fc7} features. Such concatenated representation is then used as an input to the \textit{fc8} layer that predicts important objects.  Note that we do not concatenate $X,Y$ grids with the input image so that we could preserve the original structure in the early layers of a VGG network, and use the VGG weights as an initialization. 


%

%
%
%


\subsection{Alternating Cross-Pathway Supervision}
\label{refinement}


We now describe our alternating cross-pathway supervision scheme, which is implemented via a synergistic interplay between the spatial and the visual pathways, and with a segmentation agent in between these two pathways. 



\textbf{Initial Round.} In the initial round, we want the visual pathway to predict important objects based on ``what'' they look like. It should learn to do so from the important object segmentation masks provided by an MCG projection step. These initial segmentation masks are constructed based on our guesses ``where'' important objects might appear in the first-person image.

Formally, we are given a batch of unlabeled first-person RGB images, which we denote as $B \in \mathbb{R}^{N \times C \times H \times W}$, where $N$ depicts a batch size, $H$ and $W$ refer to the height and width of an image, and $C$ refers to the number of channels ($C=3$ for RGB images). Then, let  $G \in \mathbb{R}^{N \times H \times W}$  denote images with a Gaussian placed around a spatial important object prior location ($0.6W,0.75H$). 

Furthermore let $h$ denote the MCG projection function that takes two inputs: 1) a coarse important object mask $A$, and 2) MCG regions $R$, and outputs a candidate important object segmentation mask $h(A,R)$. 

Finally, let $f(B)  \in \mathbb{R}^{N \times H \times W}$ depict the output of the visual pathway that takes a batch of first-person images as its input and outputs a per-pixel important objects map for every image in the batch. Then the cross-entropy loss that we minimize during the initial round is:

\vspace{-0.35cm}
\begin{equation*}
\begin{split}
L= - & \sum_{i=1}^N \sum_{j=1}^{H\times W} \Big[ h_j(G^{(i)},R^{(i)}) \log{(f_j(B^{(i)}))} \\
& + (1-h_j(G^{(i)},R^{(i)})) \log{(1-f_j(B^{(i)}))} \Big]
\end{split}
\end{equation*}




\textbf{V2S Round.} During the V2S (Visual to Spatial) round, given the important object masks based on ``what'' they look like,  we want spatial pathway to find image segments in the first-person image ``where'' such important objects appear.



Formally, let $g(B,X,Y)  \in \mathbb{R}^{N \times H \times W}$ depict the output of the spatial pathway, where in this case $X,Y$ denote a batch of normalized coordinate grids (each with dimensions $N \times H \times W$ ). Then the cross-entropy loss that we minimize during the V2S round is:


\vspace{-0.35cm}
\begin{equation*}
\begin{split}
L=& - \sum_{i=1}^N \sum_{j=1}^{H\times W} \Big[ h_j(f(B^{(i)}),R^{(i)}) \log{(g_j(B^{(i)},X^{(i)},Y^{(i)}))} \\
 & + (1-h_j(f(B^{(i)}),R^{(i)})) \log{(1-g_j(B^{(i)},X^{(i)},Y^{(i)}))} \Big]
\end{split}
\end{equation*}



\textbf{S2V Round.} In the S2V (Spatial to Visual) round, the visual pathway receives important object masks from the spatial pathway. Then, based on the spatial pathway's predictions  ``where'' an important object is, the visual pathway tries to learn ``what'' those important objects look like. The cross-entropy loss function that we minimize during the S2V round is:

\vspace{-0.35cm}
\begin{equation*}
\begin{split}
L=& - \sum_{i=1}^N \sum_{j=1}^{H\times W} \Big[ h_j(g(B^{(i)},X^{(i)},Y^{(i)}),R^{(i)}) \log{(f_j(B^{(i)}))} \\
& + (1-h_j(g(B^{(i)},X^{(i)},Y^{(i)}),R^{(i)})) \log{(1-f_j(B^{(i)}))} \Big]
\end{split}
\end{equation*}

\captionsetup{labelformat=empty}
\captionsetup[figure]{skip=5pt}

\begin{figure}
\centering

\myfigurethreecol{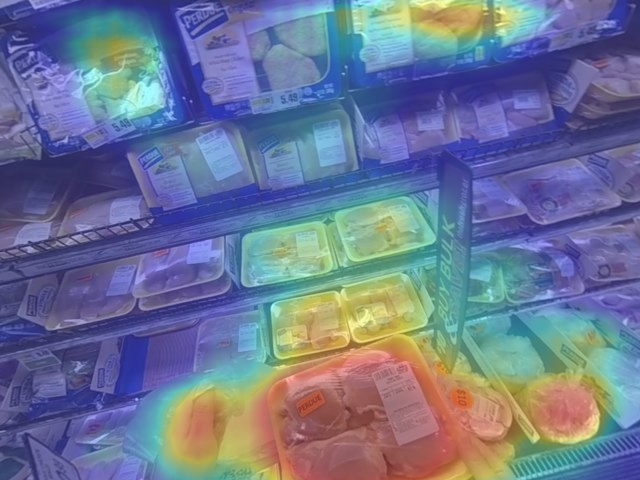}
\myfigurethreecol{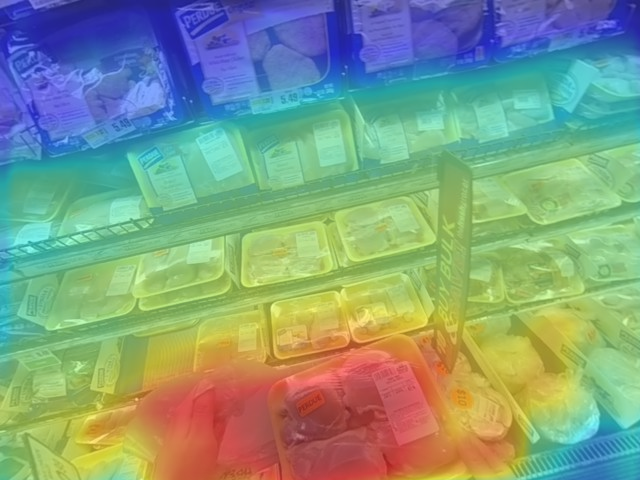}
\myfigurethreecol{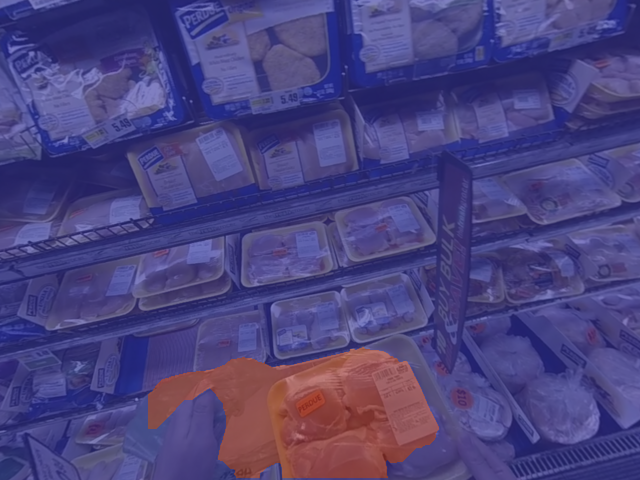}

\myfigurethreecol{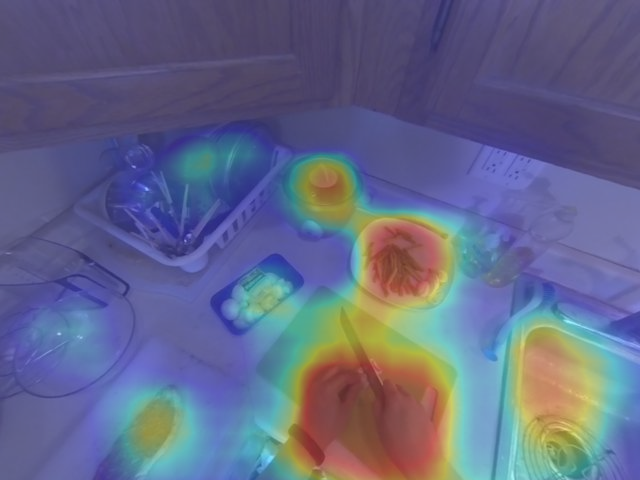}
\myfigurethreecol{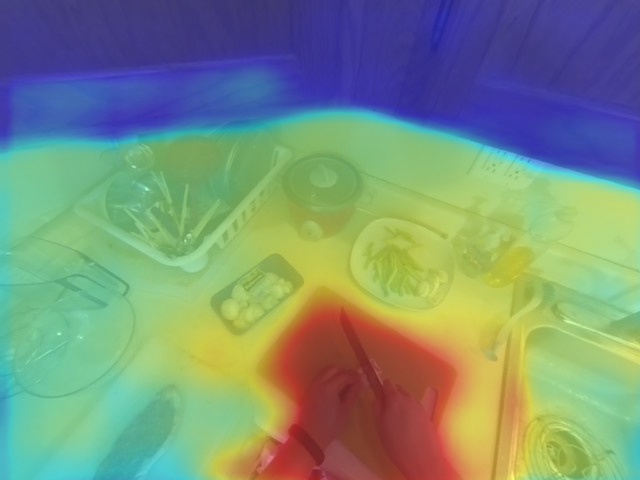}
\myfigurethreecol{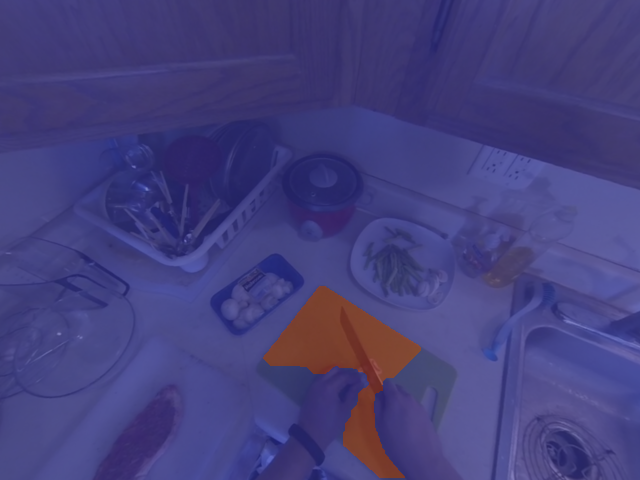}

\myfigurethreecolcaption{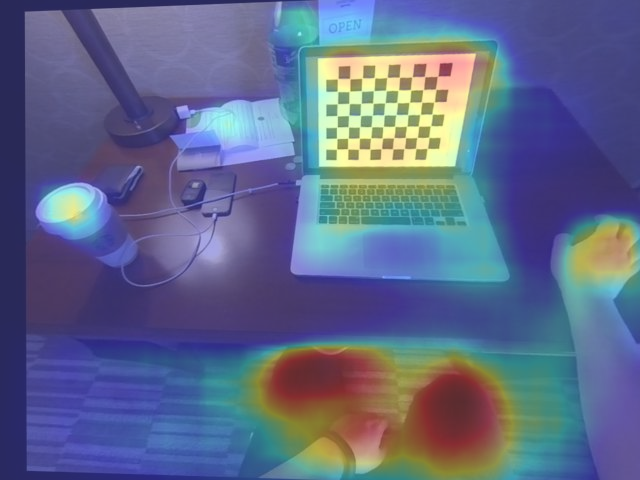}{EgoNet~\cite{DBLP:journals/corr/BertasiusPYS16}}
\myfigurethreecolcaption{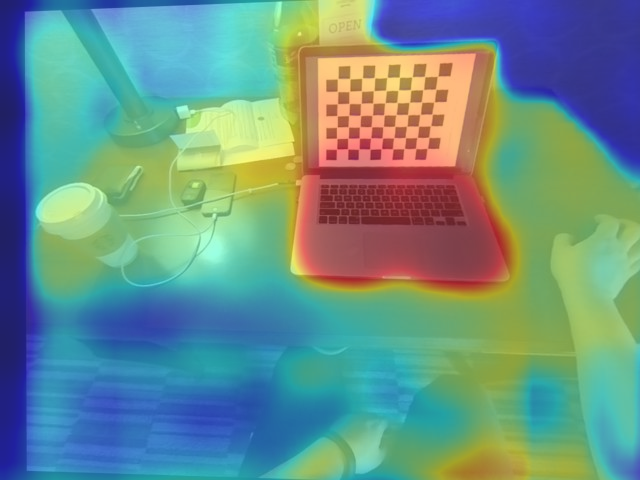}{VSN}
\myfigurethreecolcaption{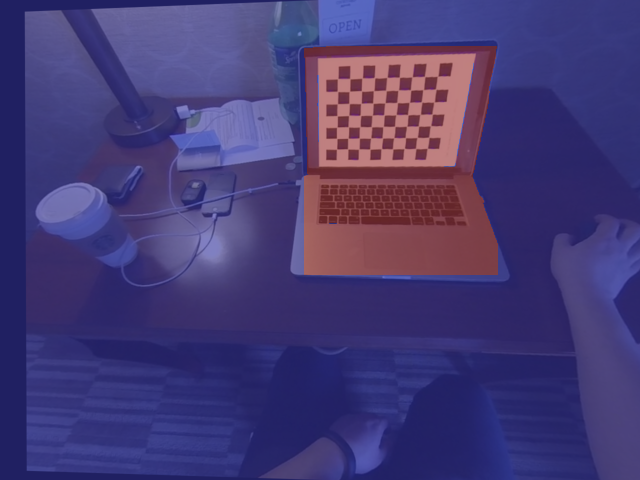}{Ground Truth}

\captionsetup{labelformat=default}
\setcounter{figure}{2}
    \caption{The qualitative important object predictions results. Despite not using any importance labels during training, our VSN correctly recognizes and localizes important objects in all three cases.\vspace{-0.5cm}}
    \label{ao_results}
\end{figure}

\captionsetup{labelformat=default}
\captionsetup[figure]{skip=10pt}

\textbf{Alternation.} We alternate our cross-pathway supervision process between the V2S and S2V rounds until there is no significant change in performance (3-4 rounds). Such an alternating learning scheme is beneficial because different visual/spatial feature inputs to the two pathways, force each pathway to maintain focus on objects that exhibit different spatial/visual characteristics.  For instance, the spatial pathway can focus on objects that are at the same spatial location, but exhibit different visual features.  In contrast, the visual pathway is able to focus on the objects that look similar but are at different locations. Such an alternation between the two pathways provides diversity to our learning scheme, which we empirically show to be beneficial.

\subsection{Using Extra Unlabeled Data for Training}
\label{extra_data}


We note that unlike supervised methods that use manually annotated importance labels, we use unlabeled data, which leads to a much harder learning task. We compensate the lack of importance labels with large amounts of unlabeled data, a strategy, which was also used by an unsupervised edge detector~\cite{Li_2016_CVPR}. For all of our experiments, we train our VSN on the combined datasets of (1) first-person important object RGBD~\cite{DBLP:journals/corr/BertasiusPYS16}, (2) GTEA Gaze+~\cite{Li_2015_CVPR}, and (3) five relevant first-person videos downloaded from YouTube (without using the labels even if they exist). We note that using more unlabeled data to train our model is essential for achieving the results that are competitive with the supervised methods' performance.

We point out that our method's ability to use unlabeled data for training is a big advantage in comparison to the supervised methods. The performance of CNNs typically improves with more training data, and unlabeled data is easy and cheap to obtain. In comparison, getting labeled data is costly and time consuming, especially if it requires per-pixel labels as in our work. 

%

\subsection{Prediction during Testing}

During testing, we average the predictions from the visual and spatial pathways. Such a prediction scheme allows each pathway to correct some of the other pathway's mistakes, and achieve a better important object prediction accuracy than any individual pathway alone would.


\subsection{Implementation Details}

For all of our experiments, we used a Caffe deep learning library~\cite{jia2014caffe}. We employed visual and spatial pathways that adapted the VGG FCN architecture~\cite{Simonyan14c}.  During training, each of the optimization rounds was set to $2000$ iterations. During those rounds one of the selected pathways was optimized to minimize the per-pixel sigmoid cross entropy loss, while the other was fixed. We performed $3$ rounds in total, which was enough to reach convergence. During the training we used a learning rate of $10^{-7}$, the momentum equal to $0.9$, the weight decay of $0.0005$, and the batch size of $15$. 




 \setlength{\tabcolsep}{3pt}
   
         \begin{table}
     \small
    \begin{center}
    \begin{tabular}{ c  | c  c  | c c ? c c |}
    \cline{2-7}
    & \multicolumn{2}{ c |}{FP-AO-RGBD}  & \multicolumn{2}{ c ?}{GTEA Gaze+} &  \multicolumn{2}{ c |}{mean}\\
    \hline
    \multicolumn{1}{| c |}{\em Method} & {\em MF} & {\em AP} & {\em MF} & {\em AP} & {\em MF} & {\em AP}\\ \hline
    	\multicolumn{1}{| c |}{VGG FCN~\cite{Simonyan14c}} & 0.166 & 0.106 & 0.325 & 0.214 & 0.246 & 0.160\\ 
	\multicolumn{1}{| c |}{GBVS~\cite{Harel07graph-basedvisual}} & 0.197 & 0.136 &  0.383 & 0.296&  0.290 &  0.216 \\
	\multicolumn{1}{| c |}{Judd~\cite{Judd_2009}} & 0.182 & 0.107 & 0.406 & 0.328 &  0.294 & 0.218 \\ 
	\multicolumn{1}{| c |}{DCL~\cite{LiYu16}} & 0.255 & 0.068 & 0.427 & 0.120 &  0.341 & 0.094 \\ 
	\multicolumn{1}{| c |}{SIOLP} & 0.278 & 0.148 & 0.416 & 0.209&  0.347 & 0.179 \\ 
	\multicolumn{1}{| c |}{Trained SIOLP$^{\ddagger}$} &  0.282 & 0.176 &  0.446 & 0.351&  0.364 &  0.264 \\ 
	\multicolumn{1}{| c |}{FP-MCG~\cite{APBMM2014}$^{\ddagger}$} & 0.317 & 0.187 & 0.447 &  0.361&  0.382 & 
 0.274 \\ 
 	\multicolumn{1}{| c |}{DeepLab~\cite{DBLP:journals/corr/ChenPKMY14}$^{\ddagger}$} & 0.370 & 0.266 &  0.472 & 0.390&  0.421 & 0.328 \\ 
 	 
 	\multicolumn{1}{| c |}{EgoNet~\cite{DBLP:journals/corr/BertasiusPYS16}$^{\ddagger}$} & 0.396 & 0.313 & \bf 0.536 & 0.449 & \bf 0.466 & 0.381 \\ 
	\multicolumn{1}{| c |}{\bf VSN} & \bf 0.421 & \bf 0.316 & 0.482 & \bf 0.472 & 0.452 & \bf 0.394 \\  \Xhline{2\arrayrulewidth}
 	\multicolumn{1}{| c |}{\bf VSN+EgoNet$^{\ddagger}$} & \bf 0.455 & \bf 0.382 &  \bf 0.588 &  \bf 0.604 &  \bf 0.522 & \bf 0.493 \\ \hline
    \end{tabular}
    \end{center}\vspace{-.3cm}
     \caption{The quantitative important object prediction results on the first-person important object RGBD and GTEA Gaze+ datasets according to the max F-score (MF) and average precision (AP) metrics. Our results indicate that even without using important object labels our VSN achieves similar or even better results than the supervised baselines. Supervised methods are marked with $^{\ddagger}$.\vspace{-0.5cm}}
    \label{ao_results_table}
   \end{table}



\section{Experimental Results}

In this section, we present quantitative and qualitative results of our VSN method. We test our method on two first-person datasets, that have per pixel important object annotations: (1) First-Person Important Object RGBD~\cite{DBLP:journals/corr/BertasiusPYS16}, and (2) GTEA Gaze+~\cite{Li_2015_CVPR} datasets. Even though both datasets have annotated importance labels, they are quite different. GTEA Gaze+ dataset captures the activities of cooking different meals, and thus there is less variation in the scene and the activity itself. In comparison, the first-person important object RGBD dataset is smaller but captures people doing seven different activities in pretty different scenes, which makes the dataset more diverse and slightly more challenging. The First-Person Important Object RGBD dataset has $4247$ annotated examples from seven video sequences, whereas for the GTEA Gaze+ dataset we use $6332$ images from $22$ different sequences.

We evaluate the important object detection accuracy using max F-score (MF), and average precision (AP), which are obtained by thresholding the probabilistic important object maps at small intervals and computing a precision and recall curve against the ground-truth important objects.


\begin{figure*}[t]
\centering

%
%

\subfigure[Spatial pathway performing better than the visual pathway]{\label{good_spatial_fig}\includegraphics[height=0.17\textheight]{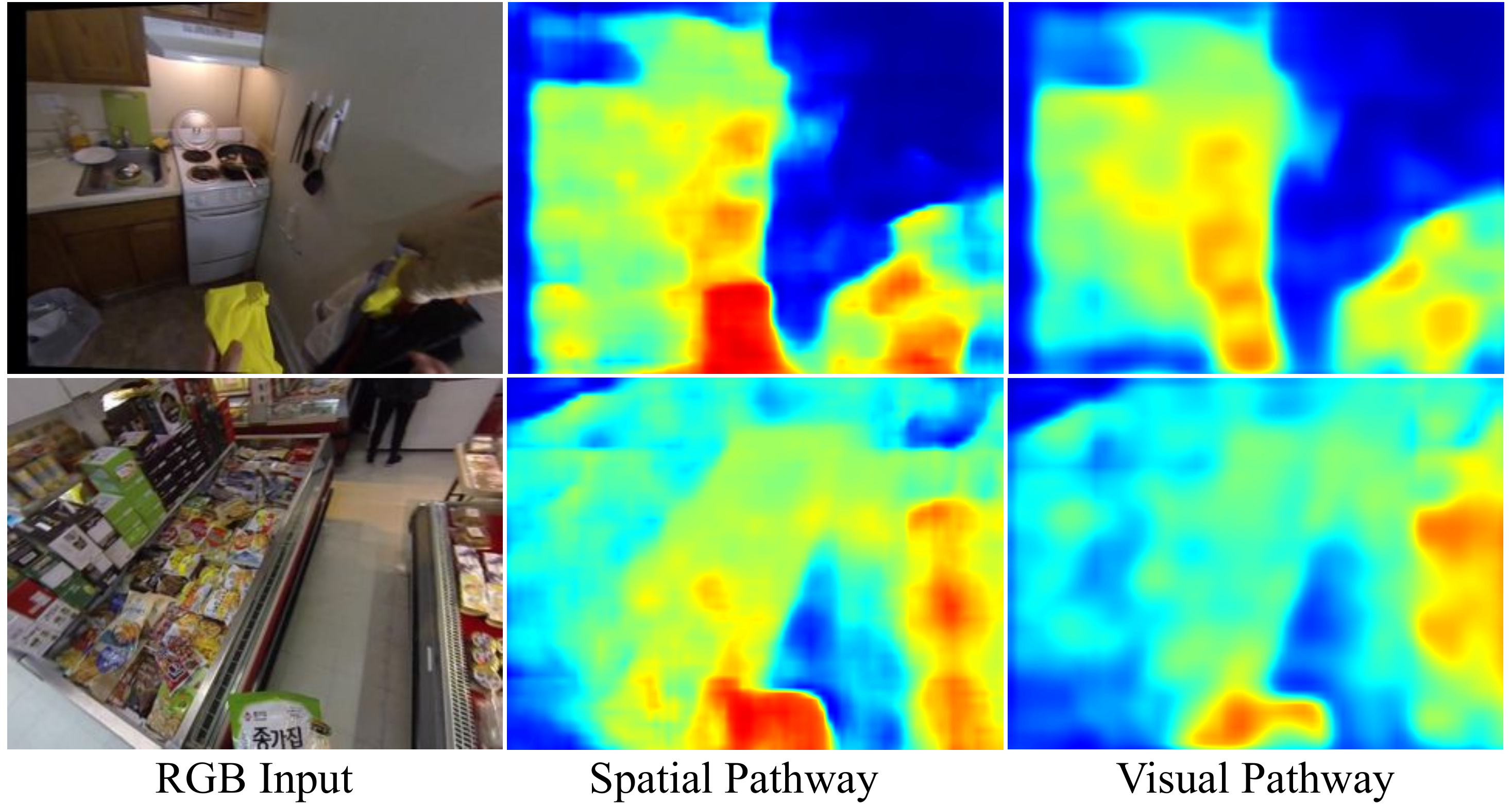}}~
\subfigure[Visual pathway performing better than the spatial pathway]{\label{bad_spatial_fig}\includegraphics[height=0.17\textheight]{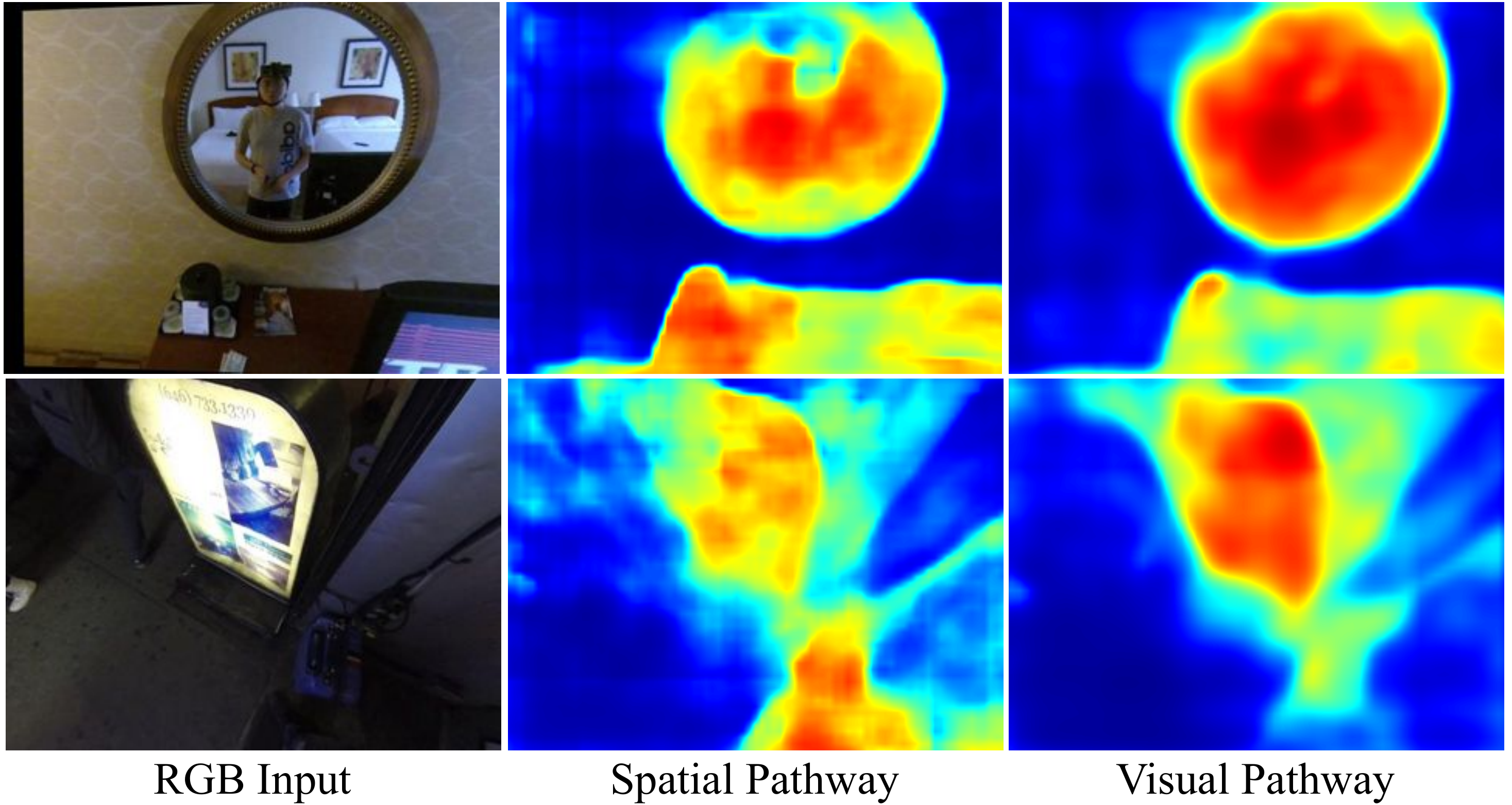}}~
\vspace{-0.4cm}
\caption{A figure illustrating a qualitative important object prediction comparison between the visual and spatial pathways (best viewed in color). Subfigure~\ref{good_spatial_fig} illustrates instances where the spatial pathway's reliance on location features is beneficial: it detects small and partially occluded important objects, which the visual pathway fails to detect accurately. The Subfigure~\ref{bad_spatial_fig} shows instances where the spatial pathway's reliance on location features leads to incorrect results: it falsely marks regions in the first-person image as important objects just because they appear at a certain location in the first-person image. In contrast, the visual pathway correctly predicts important objects in those instances.\vspace{-0.4cm}}


    \label{spatial_vs_visual_preds}
\end{figure*}

As our baselines we use a collection of the methods that were recently shown to perform well on this task as well as some of our own baselines. EgoNet~\cite{DBLP:journals/corr/BertasiusPYS16} is a two-stream network that incorporates appearance and $3D$ cues to detect important objects. We also include a DeepLab~\cite{DBLP:journals/corr/ChenPKMY14}  system, which we train for the important object detection task. Additionally, we incorporate a MCG~\cite{APBMM2014} method trained for first-person important object detection (FP-MCG). Furthermore, we include three popular visual saliency methods: (1) Judd~\cite{Judd_2009}, (2) GBVS~\cite{Harel07graph-basedvisual}, and (3) Deep Contrast Saliency method~\cite{LiYu16}. Additionally, we also evaluate the results achieved by (1) a spatial important object location prior (SIOLP), and (2) a spatial important object location prior that was obtained by extracting it from the training data using ground-truth important object labels. Furthermore, to show that the network that we used to pretrain our VSN performs poorly by itself, we include a VGG FCN~\cite{Simonyan14c} baseline. To obtain important object predictions we simply sum up the probabilities for all $20$ predicted Pascal VOC classes. Finally, to show that the predictions of our VSN method are highly complementary to the best performing EgoNet method's predictions, we combine these two methods via averaging, and demonstrate that for each dataset VSN significantly improves EgoNet's results. 

We also note that Judd~\cite{Judd_2009}, GBVS~\cite{Harel07graph-basedvisual}, DCL~\cite{LiYu16}, SIOLP, VGG-FCN, and our VSN methods do not use any important object annotations. All the other methods are trained using the manually annotated important object labels. We also note that all the FCN baselines (VGG-FCN, DeepLab, EgoNet and VSN) were pretrained for semantic segmentation under the same conditions.

We used publicly available implementations of VGG-FCN, GBVS, Judd, FP-MCG~\cite{APBMM2014}, and DeepLab~\cite{DBLP:journals/corr/ChenPKMY14} and trained and evaluated all these baselines ourselves. We obtained the results for EgoNet from the technical report in~\cite{DBLP:journals/corr/BertasiusPYS16}. To the best of our knowledge EgoNet is currently the best performing method in this task, and thus, to compare to the most recent and best performing system, we adopted the evaluation procedure from~\cite{DBLP:journals/corr/BertasiusPYS16}. 

Our evaluations provide evidence for several conclusions. In Subsections~\ref{rgbd_exp},~\ref{gtea_exp}, we show that despite \textbf{not} using any important object labels our VSN achieves results similar or even better than the supervised methods do.  Furthermore, In Subsection~\ref{abl_exp}, we provide a few ablation experiments, which show that 1) using both visual and spatial pathways is beneficial, 2) the location of an important object spatial prior is important, and that 3) using more unlabeled training data leads to better results. 
%
%
%
%



\subsection{Results on Important Object RGBD Dataset}
\label{rgbd_exp}

In Table~\ref{ao_results_table}, we present important object detection results on the First-Person Important Object RGBD dataset~\cite{DBLP:journals/corr/BertasiusPYS16}, averaged over $7$ video sequences from different activities. The results indicate that our VSN achieves the best per-class mean MF and AP scores. These results may seem surprising, because unlike EgoNet and all the other supervised baselines, our VSN does not use any important object annotations. However, VSN uses a larger amount of \textbf{unlabeled} data for its training, which leads to a better performance.




We also note that the VGG-FCN, which we use as an initialization for both of our VSN pathways, achieves the worst performance among all the baselines, which suggests that predicting $20$ Pascal VOC classes alone is not enough to achieve a good performance on the important object detection task. We also point out that combining VSN and EgoNet predictions, leads to a greatly improved accuracy according to both metrics, which implies that both methods learn complementary important object information. 

In Figure~\ref{ao_results}, we also compare qualitative important object detection results of our VSN and a supervised EgoNet model. We show that unlike EgoNet, our VSN correctly detects and segments important objects in all three cases. 





\subsection{Results on GTEA Gaze+ Dataset}
\label{gtea_exp}

In Table~\ref{ao_results_table}, we present MF and AP important object detection results on the GTEA Gaze+ dataset~\cite{Li_2015_CVPR} averaged over $22$ videos. The results indicate that our VSN outperforms all the other methods according to AP metric, and is outperformed only by EgoNet according to the MF metric. We also note that just like with the previous dataset, combining VSN and EgoNet predictions leads to a dramatic accuracy boost according to both metrics. 

\subsection{Ablation Experiments} 
\label{abl_exp}

\textbf{The Need for Spatial and Visual Pathways.} One may notice that the spatial pathway is a more powerful version of a visual pathway since it can use both spatial and visual cues to predict important objects. Therefore, a natural question is whether we need a visual pathway at all.

\begin{figure}[t]
\begin{center}
   \includegraphics[width=0.75\linewidth]{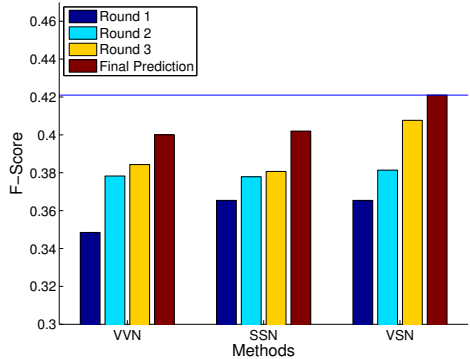}
\end{center}
\vspace{-0.4cm}
   \caption{Our results demonstrate that using a visual and a spatial pathway (VSN) yields better important object detection accuracy than using either two visual (VVN) or two spatial pathways (SSN).\vspace{-0.5cm}}
\label{fig:acc_bar}
\end{figure}

To answer this question we quantitatively compare our approach to the baselines that use either two visual pathways (VVN) or two spatial pathways (SSN). We present these results in Figure~\ref{fig:acc_bar}, where we show that our VSN method achieves $0.421$ MF accuracy,whereas the VVN and SSN baselines yield $0.402$ and $0.400$ MF accuracy respectively, suggesting that having a visual and a spatial pathway in the network is beneficial.

In Figure~\ref{spatial_vs_visual_preds}, we also present a few qualitative comparisons between the predictions from the visual and spatial pathways. In Subfigure~\ref{good_spatial_fig}, we illustrate instances where the spatial pathway's reliance on location features is beneficial: unlike the visual pathway, it is able to detect small and partially occluded important objects because they appear at a certain location. However, in Subfigure~\ref{bad_spatial_fig}, we present instances where the spatial pathway's reliance on location features leads to incorrect results: it falsely marks regions in the first-person image as important objects just because those regions appear at a certain location in an image. In contrast, in those cases, the visual pathway correctly predicts important objects because it makes the predictions based on ``what'' those objects look like.

Thus, these qualitative and quantitative results suggest that the spatial and visual pathways can complement each other, and thus, having both of them is beneficial.

\textbf{Selecting a Spatial Important Object Prior.} The initial selection of a location prior is critical to the success of our method. To validate its importance, we run several experiments on important object RGBD dataset with different location priors. First, we experiment with a location $(0.5W,0.5H)$ where $H$ and $W$ are the height and the width of an image. This is a center prior commonly used by first-person methods. Using this location prior, the pathway that was trained the last yields $0.22$ MF score, suggesting that in this case, a center location does not capture important objects well. In comparison, the pathway that was trained the last in our original model (i.e. the location prior $(0.6W, 0.75H)$) yields $0.407$ MF.

Given how important the selection of a location prior is, we need a principled way to select it. To do this we propose to utilize a generic hand detector or an unsupervised visual saliency detector on our dataset (\textbf{not}  trained on our dataset). Then, for each image we can compute a weighted average of $XY$ locations in the image (weighted by the hand detector probabilities), and then compute an average of these weighted average locations across the entire dataset.  

We report that when applied on an important object RGBD dataset, such a scheme yields a location prior of $(0.542W,0.713H)$. Training the VSN using this spatial prior then yields almost equivalent results as our original model that uses $(0.6W,0.75H)$ location prior. We also note that simply detecting hands is not enough to detect important objects.  A baseline that detects all objects and hands in the scene, and then uses objects that are closest to the hands as important object predictions yields $0.259$ MF.




\textbf{Importance of Unlabeled Training Data Size.} Finally, we also want to verify that using more unlabeled training data leads to better results. To do this, we train VSN on the same amount of \textbf{unlabeled} data, as there is labeled data used by the supervised methods ($4247$ samples). We report that using less unlabeled data leads to $0.316$ MF, which is substantially lower than our original model. 

\section{Conclusions}

In this work, we propose to detect important objects from unlabeled first-person images by formulating our problem as an interplay between the 1) recognition and 2) segmentation agents. To do this, we integrate these two agents inside an alternating cross-pathway supervision scheme of our proposed Visual-Spatial Network (VSN). The MCG projection scheme (a segmentation agent) proposes important object segmentation masks, whereas the spatial and visual pathways (recognition agents) use these masks as a supervisory signal to predict important object masks based on visual semantics and spatial features. We demonstrate the effectiveness of such scheme by showing that it achieves similar or even better results than the supervised methods.





We believe that in the future, our method could be extended to other tasks such as first-person activity recognition, or egocentric video summarization. Furthermore,  our method's ability to learn without manually annotated labels could be used to learn from large-scale unlabeled first-person datasets on the web, and in the long run, replace the supervised methods, which are constrained by the amount of available annotated data.





\bibliographystyle{plain}
\footnotesize{
\bibliography{gb_bibliography_v2,bib_hs_v2}}

\begin{thebibliography}{10}

\bibitem{APBMM2014}
P.~Arbel\'{a}ez, J.~Pont-Tuset, J.~Barron, F.~Marques, and J.~Malik.
\newblock Multiscale combinatorial grouping.
\newblock In {\em CVPR}, 2014.

\bibitem{Bearman16}
Amy Bearman, Olga Russakovsky, Vittorio Ferrari, and Li~Fei-Fei.
\newblock {What's the Point: Semantic Segmentation with Point Supervision}.
\newblock {\em ECCV}, 2016.

\bibitem{DBLP:journals/corr/BertasiusPYS16}
Gedas Bertasius, Hyun~Soo Park, Stella~X. Yu, and Jianbo Shi.
\newblock First-person action-object detection with egonet.
\newblock In {\em Proceedings of Robotics: Science and Systems}, July 2017.

\bibitem{DBLP:journals/corr/ChenPKMY14}
Liang{-}Chieh Chen, George Papandreou, Iasonas Kokkinos, Kevin Murphy, and
  Alan~L. Yuille.
\newblock Semantic image segmentation with deep convolutional nets and fully
  connected crfs.
\newblock In {\em ICLR}, 2015.

\bibitem{DBLP:journals/corr/DaiH015}
Jifeng Dai, Kaiming He, and Jian Sun.
\newblock Boxsup: Exploiting bounding boxes to supervise convolutional networks
  for semantic segmentation.
\newblock {\em CoRR}, 2015.

\bibitem{imagenet_cvpr09}
J.~Deng, W.~Dong, R.~Socher, L.-J. Li, K.~Li, and L.~Fei-Fei.
\newblock {ImageNet: A Large-Scale Hierarchical Image Database}.
\newblock In {\em CVPR09}, 2009.

\bibitem{Fathi:2011:UEA:2355573.2356302}
Alireza Fathi, Ali Farhadi, and James~M. Rehg.
\newblock Understanding egocentric activities.
\newblock In {\em ICCV}.

\bibitem{conf/cvpr/FathiRR11}
Alireza Fathi, Xiaofeng Ren, and James~M. Rehg.
\newblock Learning to recognize objects in egocentric activities.
\newblock In {\em CVPR}, pages 3281--3288. IEEE Computer Society, 2011.

\bibitem{Harel07graph-basedvisual}
Jonathan Harel, Christof Koch, and Pietro Perona.
\newblock Graph-based visual saliency.
\newblock In {\em NIPS}, 2007.

\bibitem{jia2014caffe}
Yangqing Jia, Evan Shelhamer, Jeff Donahue, Sergey Karayev, Jonathan Long, Ross
  Girshick, Sergio Guadarrama, and Trevor Darrell.
\newblock Caffe: Convolutional architecture for fast feature embedding.
\newblock {\em arXiv:1408.5093}, 2014.

\bibitem{Judd_2009}
Tilke Judd, Krista Ehinger, Fr{\'e}do Durand, and Antonio Torralba.
\newblock Learning to predict where humans look.
\newblock In {\em IEEE International Conference on Computer Vision (ICCV)},
  2009.

\bibitem{DBLP:journals/ijcv/LeeG15}
Yong~Jae Lee and Kristen Grauman.
\newblock Predicting important objects for egocentric video summarization.
\newblock {\em IJCV}, 2015.

\bibitem{LiYu16}
G.~Li and Y.~Yu.
\newblock Deep contrast learning for salient object detection.
\newblock In {\em IEEE Conference on Computer Vision and Pattern Recognition
  (CVPR)}, pages 478--487, June 2016.

\bibitem{Li_2016_CVPR}
Yin Li, Manohar Paluri, James~M. Rehg, and Piotr Dollar.
\newblock Unsupervised learning of edges.
\newblock In {\em The IEEE Conference on Computer Vision and Pattern
  Recognition (CVPR)}, June 2016.

\bibitem{Li_2015_CVPR}
Yin Li, Zhefan Ye, and James~M. Rehg.
\newblock Delving into egocentric actions.
\newblock In {\em CVPR}.

\bibitem{Lin_2016_CVPR}
Di~Lin, Jifeng Dai, Jiaya Jia, Kaiming He, and Jian Sun.
\newblock Scribblesup: Scribble-supervised convolutional networks for semantic
  segmentation.
\newblock In {\em The IEEE Conference on Computer Vision and Pattern
  Recognition (CVPR)}, June 2016.

\bibitem{502}
Tsung-Yi Lin, Michael Maire, Serge Belongie, James Hays, Pietro Perona, Deva
  Ramanan, Piotr Doll{\'a}r, and C.~Lawrence Zitnick.
\newblock Microsoft coco: Common objects in context.
\newblock In {\em European Conference on Computer Vision (ECCV)}, Z{\"u}rich,
  September 2014.

\bibitem{DBLP:journals/corr/LongSD14}
Jonathan Long, Evan Shelhamer, and Trevor Darrell.
\newblock Fully convolutional networks for semantic segmentation.
\newblock {\em CoRR}, 2014.

\bibitem{Lu:2013:SSE:2514950.2516026}
Zheng Lu and Kristen Grauman.
\newblock Story-driven summarization for egocentric video.
\newblock In {\em CVPR}, 2013.

\bibitem{ma2016going}
Kris~Kitani Minghuang~Ma.
\newblock Going deeper into first-person activity recognition.
\newblock In {\em Conference on Computer Vision and Pattern Recognition
  (CVPR)}, 2016.

\bibitem{papandreou15weak}
George Papandreou, Liang-Chieh Chen, Kevin Murphy, and Alan~L Yuille.
\newblock Weakly- and semi-supervised learning of a dcnn for semantic image
  segmentation.
\newblock {\em arxiv:1502.02734}, 2015.

\bibitem{pathakICCV15ccnn}
Deepak Pathak, Philipp Kr\"ahenb\"uhl, and Trevor Darrell.
\newblock Constrained convolutional neural networks for weakly supervised
  segmentation.
\newblock In {\em ICCV}, 2015.

\bibitem{pathakICLR15}
Deepak Pathak, Evan Shelhamer, Jonathan Long, and Trevor Darrell.
\newblock Fully convolutional multi-class multiple instance learning.
\newblock In {\em ICLR Workshop}, 2015.

\bibitem{pinheiro:2015a}
P.~H.~O. Pinheiro and R.~Collobert.
\newblock From image-level to pixel-level labeling with convolutional networks.
\newblock In {\em Conference on Computer Vision and Pattern Recognition
  (CVPR)}, 2015.

\bibitem{PirsiavashR_CVPR_2012_1}
Hamed Pirsiavash and Deva Ramanan.
\newblock Detecting activities of daily living in first-person camera views.
\newblock In {\em CVPR}, 2012.

\bibitem{conf/cvpr/RenG10}
Xiaofeng Ren and Chunhui Gu.
\newblock Figure-ground segmentation improves handled object recognition in
  egocentric video.
\newblock In {\em CVPR}, 2010.

\bibitem{Simonyan14c}
K.~Simonyan and A.~Zisserman.
\newblock Very deep convolutional networks for large-scale image recognition.
\newblock {\em arXiv:1409.1556}, 2014.

\bibitem{xu_cvpr2015}
Jia Xu, Alexander~G. Schwing, and Raquel Urtasun.
\newblock Learning to segment under various forms of weak supervision.
\newblock In {\em Proc. CVPR}, 2015.

\end{thebibliography}

\end{document}